\documentclass[twoside,leqno,twocolumn]{article} 
\usepackage{ltexpprt} 
\usepackage{algorithm}
\usepackage{algpseudocode}
\usepackage{amsmath,amssymb,amsfonts}
\usepackage{graphicx}
\graphicspath{{Figures/}}
\usepackage[labelformat=simple]{subcaption}

\usepackage{mathtools}
\newtheorem{definition}{Definition}[section]
\usepackage{color}
\usepackage{array,multirow}
\DeclareMathOperator*{\argmax}{arg\,max}

\makeatletter
\def\bstctlcite{\@ifnextchar[{\@bstctlcite}{\@bstctlcite[@auxout]}}
\def\@bstctlcite[#1]#2{\@bsphack
	\@for\@citeb:=#2\do{%
		\edef\@citeb{\expandafter\@firstofone\@citeb}%
		\if@filesw\immediate\write\csname #1\endcsname{\string\citation{\@citeb}}\fi}%
	\@esphack}
\makeatother

\begin{document}
\bstctlcite{IEEEexample:BSTcontrol}
\algnewcommand{\LeftComment}[1]{\Statex \(\triangleright\) \textcolor{blue}{#1}}

\title{\Large Mining Sub-Interval Relationships In Time Series Data}
\author{Saurabh Agrawal \thanks{Department of Computer Science, University of Minnesota} \\
\and Saurabh Verma \footnotemark[1] \\
\and Gowtham Atluri \thanks{Department of EECS, University of Cincinnati} \\
\and Anuj Karpatne \footnotemark[1]\\
\and Stefan Liess \thanks{Department of Earth Sciences, University of Minnesota} \\
\and Angus Macdonald III \thanks{Department of Psychology, University of Minnesota} \\
\and Snigdhansu Chatterjee \thanks{School of Statistics, University of Minnesota}\\
\and Vipin Kumar \footnotemark[1]\\
}
\date{}
\maketitle

\maketitle


\begin{abstract}  \small\baselineskip=9pt
Time-series data is being increasingly collected and studied in several areas such as neuroscience, climate science, transportation, and social media. Discovery of complex patterns of relationships between individual time-series, using data-driven approaches can improve our understanding of real-world systems. While traditional approaches typically study relationships between two entire time series, many interesting relationships in real-world applications exist in small sub-intervals of time while remaining absent or feeble during other sub-intervals. In this paper, we define the notion of a sub-interval relationship (SIR) to capture interactions between two time series that are prominent only in certain sub-intervals of time. We propose a novel and efficient approach to find most interesting SIR in a pair of time series. We evaluate our proposed approach on two real-world datasets from climate science and neuroscience domain and demonstrated the scalability and computational efficiency of our proposed approach. We further evaluated our discovered SIRs based on a randomization based procedure. Our results indicated the existence of several such relationships that are statistically significant, some of which were also found to have physical interpretation. 

\end{abstract}

\section{Introduction}
    Time series data that consists of a sequence of observations collected at regular intervals of time is ubiquitous to many real-world applications. Mining relationships in time series data is of immense interest to several disciplines such as neuroscience, climate science, and transportation. For example, in climate science, relationships are studied between time series of physical variables such as Sea Level Pressure, temperature, etc., observed at different locations on the globe. Such relationships, commonly known as `teleconnections' capture the underlying processes of the Earth's climate system \cite{kawale2013graph}. Similarly, in neuroscience, relationships are studied between activities recorded at different regions of the brain over time \cite{atluri2016brain,atluri2015connectivity}. Studying such relationships can help us improve our understanding of real-world systems, which in turn, could play a crucial role in devising solutions for   problems such as mental disorders or climate change.


\begin{figure*}[h]
\centering
\begin{subfigure}[t]{0.22\textwidth}
\includegraphics[scale=0.29]{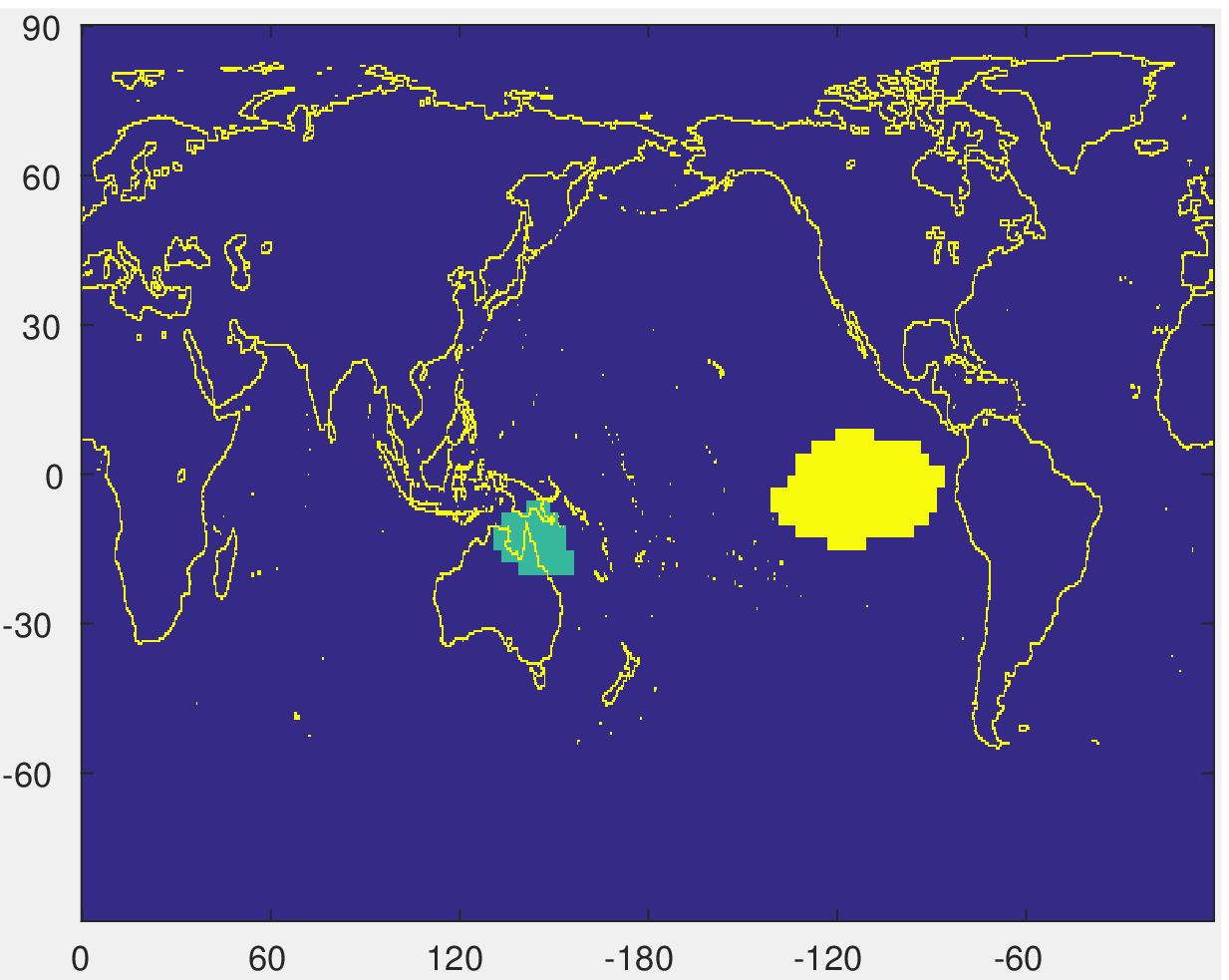}
\centering
\caption{Two ends of El-nino Southern Oscillation}
\label{fig:ENSOReg}
\end{subfigure}
\begin{subfigure}[t]{0.7\textwidth}
\includegraphics[scale=0.4]{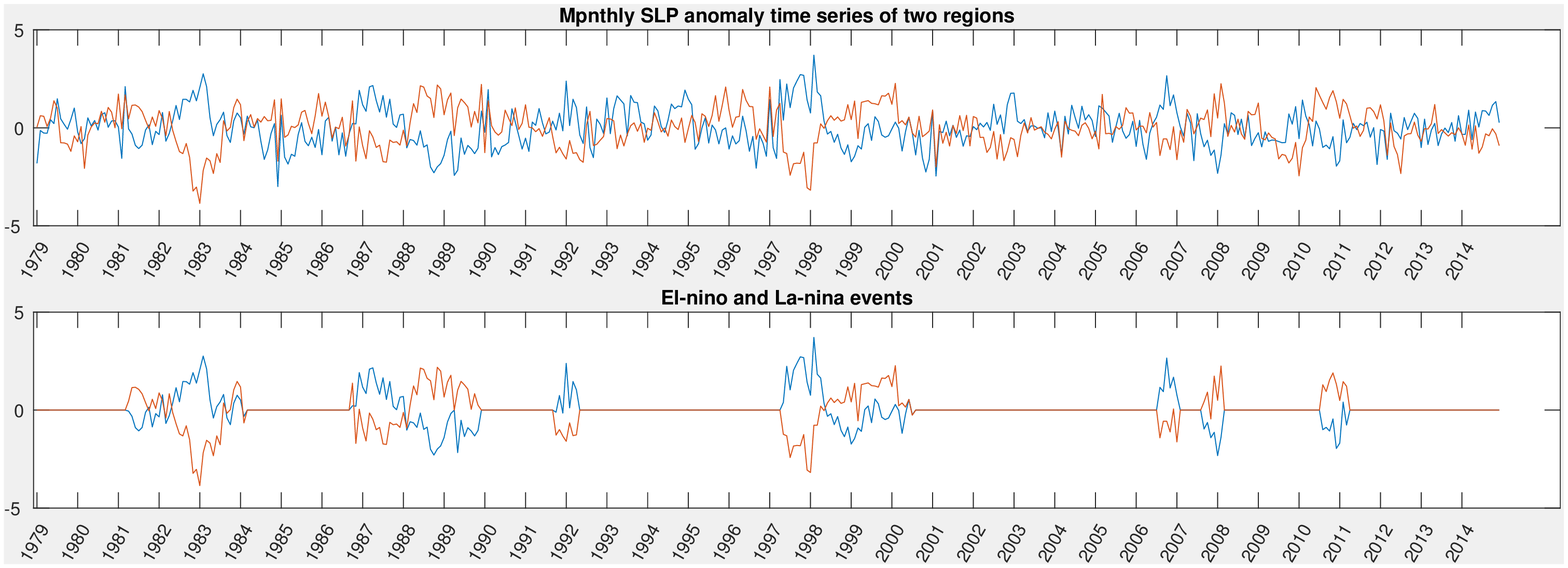}
\centering
\caption{Area-averaged Sea Level Pressure (SLP) monthly time series of two regions}
\label{fig:ENSOTs}
\end{subfigure}
\vspace{-0.5em}
\caption{An example of a Sub-Interval Relationship from climate science}
\vspace{-1.5em}
\end{figure*}

Most of the existing work on mining time series relationships assume the relation to be present for the entire duration of the two time series. However, many interesting relationships in real-world applications often are intermittent in nature, i.e., they are highly prominent only in certain sub-intervals of time and absent or occur feebly in rest of the sub-intervals. As a motivating example, consider a pair of monthly Sea Level Pressure anomaly time series during 1979-2014 in Figure~\ref{fig:ENSOTs} that are observed at two distinct regions on the globe in Figure~\ref{fig:ENSOReg}). The full-length correlation between the two time series is -0.25. However, as shown in the lower panel of Figure~\ref{fig:ENSOTs}, there exists multiple sub-intervals where the correlation between the two time series is stronger than -0.7. As we discuss later in Section~\ref{Sec:DomInsights}, this example is the outcome of a well-known climate phenomena called ENSO (El Nino Southern Oscillation) \cite{glantz2001currents}, that is characterized by negative correlations between the surface temperatures observed near Australia and Pacific Ocean \cite{glantz2001currents} and is known for impacting various weather events such as floods, droughts, and forest fires \cite{siegert2001increased,ward2014strong}. The sub-intervals shown in the lower panel correspond to the two extreme phases, `El-Nino' and `La-nina', of ENSO, when its impact on global climate is amplified. Similar examples are also known to exist in other domains such as neuroscience, \cite{atluri2014discovering} and stock market data \cite{li2016efficient}. 

Inspired by such examples, we propose to formally study sub-interval relationship (SIR) between two time-series. Studying SIRs in time series data poses several challenges. First, we need to formally define the notion of an SIR and devise necessary interestingness measures to characterize them. Second, given a pair of time series, a computationally efficient method needs to be designed in order to find all the sub-intervals that show strong relationships. Third, the validity of discovered relationships needs to be evaluated.

For a given pair of time series, we define SIR as the set of all the disjoint time intervals each of which are sufficiently long and capture strong relationships between the two time series. The constraint on the length of the interval is important to filter out very small intervals that are more likely to capture spurious signals. An SIR with longer and multiple such intervals can be treated to be more interesting. Thus, the interestingness of an SIR could be measured as the total sum-length of all   included intervals. 

The next challenge is then to obtain most interesting SIR in a given pair of time series. As we describe later in Section~\ref{Sec:DP}, the above problem can  exactly be solved by a standard dynamic programming (DP) approach. However, the time complexity of this approach is quadratic in the length of time series, which makes it computationally infeasible for long time series. To the best of our knowledge, none of the existing work in the literature solves our exact problem formulation and are therefore not applicable (discussed further in Section~\ref{Sec:RelWork}). In this work, we propose a novel and efficient approach called \textit{Partitioned Dynamic Programming}(PDP) to find the most interesting SIR in a given pair of time series. The basic idea of our approach is to partition the original problem into multiple sub-problems, each of which could be independently solved by standard dynamic programming methods. We show that our approach is guaranteed to find the optimal solution and has time complexity that is practically linear in the length of the time series. Finally, to evaluate the validity of discovered SIRs, we propose a randomization based procedure to assess their statistical significance.

\section{Definitions and Problem Formulation}\label{Sec:Def}
\subsection{Definitions and Notations}
We start with presenting some basic terms and notations that will be used throughout  the paper. \begin{enumerate}\setlength\itemsep{-0.3em}
    \item \textit{Time series}   $T$ is a sequence of $N$ observations made at consecutive and regular timestamps. 
    \item \textit{Time interval}  $[s,e]$ represents a set of consecutive timestamps $\{s,s+1,s+2,...,e-1,e\}$. Time interval with a single timestamp $t$ is given by $[t]$.
    \item \textit{Length of a time interval} $[s,e]$ represents the number of timestamps included in the interval and denoted by $l_{s,e}$. The length of single timestamp $t$ is equal to 1 and is denoted by $l_{t,t}$.
    \item \textit{Segment of a time series} $T[s,e]$ in a time interval $[s,e]$ contains all the observations of $T$ corresponding to the timestamps of given interval. 
    \item \textit{Full-length relationship between two time series} $T_1$ and $T_2$ refers to the strength of  relationship computed using all the observations of the two time series and is denoted by $rel(T_1,T_2)$. 
    \item \textit{Local relationship between two time series} $T_1$ and $T_2$ in a time interval $[s,e]$ refers to the strength of relationship computed using all the observations in the time interval and is denoted by $rel(T_1[s,e],T_2[s,e])$. For brevity, we also denote it by $rel[s,e]$. Certain relationship measures (e.g., Euclidean distance) could be computed over singleton timestamp $t$, which would be denoted by $rel[s]$. 
\end{enumerate}

Following these ideas, we present our formal definition of sub-interval relationships (SIR) between a pair of time series. 

\begin{definition}
A \textit{Sub-Interval Relationship (SIR)} between two time series $T_1$ and $T_2$ refers to the set $S$ of non-overlapping time intervals $\{[s_1,e_1],[s_2,e_2],...,[s_n,e_n]\}$ such that every interval in $S,$
    \begin{itemize}\setlength\itemsep{-0.5em}
        \item captures strong relationship, i.e. $rel[s,e] \geq \tau$ $\forall [s,e] \in S.$
        \item is of length at least  $l_{min}$, i.e. $l_{se} \geq l_{min}\forall [s,e] \in S.$
    \end{itemize}
    where $\tau$ and $l_{min}$ are user-specified thresholds.      
\end{definition}  
The choice of thresholds depends on the type of SIRs that are of interest to a user. For instance, setting higher $l_{min}$  and lower $\tau$ results in SIRs with longer intervals of mild relationships and vice-versa. 

Note that our definition of SIR is quite general and not specific to any relationship measure. Further, the above definition could also be used with distance measures (e.g. euclidean distance) except that the inequality constraint in the first bullet flips, i.e. $dist[s,e] \leq \tau$.
\vspace{-2em}
\subsection{Problem Formulation}
Our next goal is to capture the most interesting SIR for a given pair of time series. Intuitively, an SIR is likely to be more interesting if the set $S$ of selected intervals covers a large fraction of the timestamps. Therefore, we propose to measure interestingness of an SIR as the \textbf{\textit{sum-length}} ($SL$), which is equal to sum of lengths of all the selected time intervals. Then the problem require us to find the set of long and strong time-intervals with maximum sum-length. 

Formally, for a given pair of time series $(T_1,T_2)$, our goal is to determine the most interesting SIR as the set $S$ of time-intervals with maximum sum-length such that for every interval $[s,e] \in S$, $rel[s,e] \geq \tau$, and $l_{se} \geq l_{min}$.

where $l_{min}$ and $\tau$ are user-specified thresholds. In the remainder of the paper, we will refer to the set $S$ with maximum sum-length as the `\textbf{optimal set}'.


\vspace{-0.5em}

\section{Related Work}\label{Sec:RelWork}
Learning the similarity between two time series has been well studied in many different settings. However, quite surprisingly, our problem formulation has never been studied in any of the existing works. 
The most prevalent work has been done in designing various similarity measures (e.g., euclidean distance, Pearson correlation, dynamic time warping) for  analyzing full-length time series~\cite{kawale2013graph,keogh2002exact,liao2005clustering}. 
Another part of the related work goes into devising various largest common subsequence (LCS) matching problems \cite{das1997finding,chen2007spade,faloutsos1994fast}. Other related works focus on all-pairs-similarity-search and motif discovery~\cite{yeh2016matrix,zhu2016matrix} in which the fundamental problem is  to efficiently find all the subsequences of a time series that are the closest match (or match within a threshold) to a given query sequence, that represents a pattern of high interest (e.g. motif) or obtained as a subsequence from another time series. 
The key difference between LCS/motif-discovery and our problem formulation is that the former allows matching between observations of two time series at different timestamps by compromising on the computational complexity. In contrast, we restrict ourselves to lock-step relationship measures (i.e. $i^{th}$ time-point in a time series can be compared with only $i^{th}$ time-point in other time series) that can be solved much more efficiently and has direct relevance to various domains including climate science and neuroscience. 

The \emph{closest setting} related to our work was formulated in~\cite{li2013discovering,li2016efficient}, where a longest subsequence  of atleast $\delta$ threshold (overall) correlation was considered but disregards any constraints on the sub-intervals. 
Like ours, their work also stress on the importance of efficiently computing the subsequence correlation but rely heavily on bringing down the cost of \emph{constant factors}  involved in correlation computation and therefore, still bounded by $\mathcal{O}(N^2)$. A tangential amount of work can also be found in some graphical based models such as hidden-semi markov models (HsMMs) and segmental hidden markov models~\cite{liang2011improved,kim2006segmental}. However, such approaches often lack optimal guarantees due to the non-convexity of the objective function.

\vspace{-0.5em}
\section{Methodology}\label{Sec:Method}
Our problem formulation can potentially be solved by two approaches: (i) a classical approach based on the dynamic programming, and (ii) our proposed approach -- Partitioned Dynamic Programming, that is an extension of the classical dynamic programming. We now elaborate the two approaches in further details. 
\vspace{-0.5em}
\subsection{Classical Approach: Dynamic Programming}\label{Sec:DP}
The problem of finding the optimal set can be treated as a classical DP problem of weighted interval scheduling \cite{kleinberg2005algorithm} where the goal is to determine a schedule of jobs such that no two jobs conflict in time and the total sum of the weights of all the selected jobs is maximized. In our problem, we can treat every time-interval that meets the minimum strength and length criteria as a job and the interval length as the weight of the job. We could then use DP to find the set of intervals with maximum possible sum-length. 

The key condition for dynamic programming is to decompose a larger problem into sub-problems such that each sub-problem can further be  solved recursively. Let $S_{k}$ denote the optimal set to the problem spanning timestamps $[1,k]$, and let $SL_{k}$ denote the sum-length of $S_{k}$. Then the recursive relation between sum-lengths of the optimal sets of different sub-problems is given as, 
\vspace{-0.8em}
$$SL_{N} = \max\limits_{k \in [1,N]}(W[k][N]+SL_{N-1})$$



In other words, the optimal set $S_{N}$ is either exactly same as $S_{N -1}$ or can be obtained by adding the interval $[k,N]$ to the optimal set $S_{k-1}$ for one of the values of $k \in [1,N-1]$. 
By applying above recursions, the optimal sets for all the sub-problems and the original problem can be obtained as described  in Algorithm~\ref{AlgoDP}.

\begin{algorithm}[!h]
\footnotesize
  \caption{Dynamic Programming to find SIR between a pair of time series}\label{AlgoDP}
\begin{algorithmic}[1]
  \State{\textbf{Input:}Time series $T_1$,$T_2$, Parameters:$\tau,l_{min}$}
  \State{\textbf{Output:} A set $S_N$ of time intervals that form an SIR between $T_1$ and $T_2$}
  \State{$W \gets$ a $N \times N$ matrix that stores the weight of each interval as its length}
  \State{$W[i][j] = 0, \forall i,j$ such that $rel[i,j] < \tau$ or $l_{ij}<l_{min}$}
  \State{$S_{i} \gets$ the optimal set for the sub-problem spanning timestamps between 1 and $i$ initialized to $\phi$}
  
  \State{$SL_{i}\gets$ sum-lengths of $S_{i}$ initialized to zero}
  \State{$N \gets$ length of input time series}
  \For{$i = 1:N$}
   \State{$SL_{i} = \max\limits_{k \in [1,i]}(W[k][i]+SL_{k-1})$}
   \State{$k = \argmax\limits_{k \in [1,i]}(W[k][i]+SL_{k-1})$}
    \If {$SL_{i} > SL_{i-1}$}
        \State{$S_{i} \gets S_{k-1} \cup [k,i]$}
    \Else
        \State{$S_{i} \gets S_{i-1}$}
    \EndIf
  \EndFor
  \State{\textbf{return} $S_{N}$} 
\end{algorithmic}
\end{algorithm}

\begin{algorithm}[!h]
\footnotesize
  \caption{Partitioned Dynamic Programming}\label{AlgoPDP}
  \begin{algorithmic}[1]
  \Statex{\textbf{Input}:Time series $T_1$,$T_2$, Threshold on interval strength:$\tau$}
  \Statex{\textbf{Output:} SIR between $T_1$ and $T_2$}
  \Statex{\textbf{Step 1}}
  \State{$LW \gets $ All timestamps that satisfy left-weakness}
  \Statex{\textbf{Step 2}}
  \State{$RW \gets $ All timestamps that satisfy right-weakness}
  \Statex{\textbf{Step 3}}
  \State{$PoP \gets $ $LW$ AND $RW$} \Comment{Points of Partition}
  \State{Obtain all the partitions between every two adjacent points of partition}
  \State{Use Algorithm~\ref{AlgoDP} to solve the sub-problem for each partition}
  \State{$SIR(T_1,T_2) \gets $ Union set of optimal sets to all the sub-problems}
  \end{algorithmic}
\end{algorithm} 
\vspace{-1em}

\subsubsection{Time Complexity}\label{sec:DPcomplx} For any pair of time series, dynamic programming needs to solve all the $N$ sub-problems of sizes $1,2,...,N$, each of which takes $\mathcal{O}(m)$ time, where $m$ is the length of the sub-problem. Therefore, both the average-case and worst-case time complexity of dynamic programming is $O(N^2)$ in time. 

 
\subsection{Proposed Approach (Partitioned Dynamic Programming)}\label{sec:PDP}
As discussed above, the time complexity of DP is $O(N^2)$, which could be computationally infeasible for longer time series. A potential approach to reduce computational cost could be to partition the original problem into multiple sub-problems of constant sizes and solving each of them independently using DP. The optimal set to the original problem could then be obtained by taking union of the optimal sets obtained for all sub-problems. The computational cost of this approach would depend on the sizes of the different sub-problems. If the size of each sub-problem is smaller than a constant $k$, then the computational cost would be $O(\frac{N}{k}*k^2) = O(N)$, which would be faster than DP by an order of $N$. However, a key challenge in this approach is to partition the problem prudently such that no interesting interval gets fragmented across the two partitions, otherwise it could potentially get lost if its fragmented parts are not sufficiently long or strong enough to meet the user-specified thresholds.

To this end, in this section we propose a novel approach called \textit{Partitioned Dynamic Programming} (PDP) that is significantly efficient than Dynamic Programming (DP) and is guaranteed to find the optimal set. PDP follows the above idea and breaks the original problem into multiple sub-problems such that each one of them can be solved by using DP independently.
The key step in PDP is to identify safe \emph{points of partition}, where the problem could be partitioned without compromising on the optimality of the solution. However, we would need to mention that unlike DP, PDP is applicable only to those relationship measures that satisfy the following three properties: 

\noindent \textbf{Property 1:} The relationship measure could be computed over a single timestamp. \\
\textbf{Property 2:} If $rel[s,e]$ is known, then $rel[s,e+1]$ and $rel[s-1,e]$ could be computed in constant time. \\
\textbf{Property 3:} For a given pair of time series, let $[s,m]$ and $[m+1,e]$ be two adjacent time-intervals, $\alpha = min(rel[s,m]$, $ rel[m+1,e])$, and $\beta = max(rel[s,m]$, $ rel[m+1,e])$, then $\alpha \leq rel[s,e]$ $\leq$ $\beta$. \\
The above three properties are satisfied by various measures that we discuss in more detail in section~\ref{sec:Meas}.





From Property 3, it follows that an interval $[s,e]$ formed by   union of two adjacent weak intervals $[s,m]$ and $[m+1,e]$ could never be strong. Thus, a timestamp $t$ can be considered as a `point of partition' if:
\vspace{-0.5em}
\begin{enumerate}  \setlength\itemsep{-0.5em}
	  \item none of the intervals ending at $t-1$ are strong, i.e. $rel[s,t-1]< \tau \forall s \in [1,t-1]$. We refer to this condition as \textbf{left-weakness} condition.
    \item none of the intervals beginning from $t$ are \emph{strong}, i.e. $rel[t,e] < \tau \forall e \in [t,L]$. We refer to this condition as \textbf{right-weakness} condition. 
\end{enumerate}
\vspace{-0.5em}
The two conditions above ensure that all the intervals ending at $t-1$ or beginning from $t$ are weak, therefore no strong interval that subsumes $t$ could possibly exist.  Therefore, no interesting interval would be in danger of getting fragmented, if the problem is partitioned at $t$. Following this idea, we propose a partitioning scheme to find the points of partition before applying dynamic programming module to each of the partitions. 

As summarized in Algorithm~\ref{AlgoPDP}, PDP comprises of three major steps: In step 1, we find all timestamps $t$ such that they satisfy the left-weakness property. In step 2, we identify all the timestamps $t$ that satisfy the right-weakness property. Finally, in step 3, all the timestamps that satisfy both left-weakness and right-weakness are considered as the points of partition. The original problem is then partitioned at the obtained points of partition and the resulting sub-problems are solved independently using the dynamic programming module described in Algorithm~\ref{AlgoDP}. 

We next explain the two procedures used in Step 1 and Step 2 to obtain points that satisfy the left-weakness and right-weakness respectively. 

\subsubsection{Finding timestamps with left-weakness} To find timestamps with left-weakness, we perform a left-to-right scan of timestamps as summarized in Algorithm~\ref{AlgoLW}. We begin our scan from the leftmost timestamp to find the first timestamp s such that $[s]$ is strong, i.e. $rel[s] \geq \tau$. We next show that all the timestamps $\{2,...,s\} $ will satisfy left-weakness using the following lemma.
\begin{lemma}\label{lem1}
Consider a timestamp $t$ that satisfies left-weakness. If $rel[t] < \tau$, then $t+1$ would also satisfy left-weakness. (\textbf{Proof in supplementary})
\end{lemma}
%
Since there are no timestamps to the left of first timestamp, it trivially satisfies left-weakness. By recursively applying Lemma~\ref{lem1}, to timestamps $\{2,3,...,s\}$, we get each of them to satisfy left-weakness. 

We then continue our scan beyond $s$ to find the first timestamp $e$ such that $[s,e]$ is weak, i.e. $rel[s,e] < \tau$ (lines 15-21). This also means that for all the timestamps $m$ $\in$ $[s,e]$, interval $[s,m-1]$ is strong, and therefore $m$ violates left-weakness. We next show that timestamp $e+1$ satisfies the left-weakness as follows,

\begin{lemma}\label{lem2}
Consider a set of timestamps $S =\{s,s+1,...,e-1,e\}$ such that $rel[s,m] \geq \tau \forall m \in [s,e-1]$, while $rel[s,e] < \tau$. If $s$ satisfies left-weakness, then timestamp $e+1$ would also satisfy left-weakness. (\textbf{Proof in supplementary})
\end{lemma}
%

\begin{algorithm}[!t]
\footnotesize
  \caption{Find timestamps that satisfy left-weakness}\label{AlgoLW}
  \begin{algorithmic}[1]
  \Statex{\textbf{Input}:Time series $T_1$,$T_2$, Threshold on interval strength:$\tau$}
  \Statex{\textbf{Output:}A boolean array $LW$ s.t. $LW[i] = TRUE$ iff $i$ satisfies left-weakness}
  \State{$LW \gets $  N $\times$ 1 boolean array s.t. LW[i] = FALSE $\forall i \in [1,N]$}\Comment{N: length of time series}
  \State{LW[1] = TRUE}
  \State{scan finish = FALSE}
  \State{t = 1} 
  \While{NOT scan finish}
    \While{rel[t] $<\tau$}
        \State{LW[t+1] = TRUE}
        \State{t = t+1}
        \If{t$==N-1$}
            \State{scan finish = TRUE}
            \State{\textbf{break}}
        \EndIf
    \EndWhile
    
    \State{s = t}
    \While{NOT scan finish AND rel[s,t] $\geq \tau$}
        \State{t = t+1}
         \If{t$==N-1$} 
            \State{scan finish = TRUE} 
         \EndIf
    
    \EndWhile
  \EndWhile
  \State{\textbf{return} LW}
  \end{algorithmic}
\end{algorithm}

We further continue our scan and repeat the above steps (lines 6-20) to find all the timestamps that satisfy left-weakness. In summary, the above procedure essentially finds streaks of points that satisfy or violate left- weakness in a single scan. Similar procedure could be followed to find timestamps that satisfy right-weakness except that the scan would proceed leftwards starting from the rightmost timestamp. An illustration of PDP is provided in supplementary. 

\subsubsection{Time Complexity}\label{Sec:PDPcomplx}
There are three major steps in PDP. In the first step, we use Algorithm~\ref{AlgoLW} to find points that satisfy left-weakness. Notice that each timestamp is visited exactly once in the scan and under the assumption of Property 2, the computation of $rel[s,t]$ in line 15 of Algorithm~\ref{AlgoLW} takes constant time. Therefore, the time complexity of Algorithm~\ref{AlgoLW} is $O(N)$. By similar arguments, the complexity of Step 2 is also $O(N)$. Thus, the time complexity of finding points of partition is $O(N)$. 

The total time complexity of PDP is therefore $O(N)+O(NK)$, where K is the average computational cost of the sub-problems corresponding to every partition. The worst-case complexity is $O(N) + O(N^2)$, that corresponds to the cases where either no partition could be obtained or the largest partition obtained is of $O(N)$ in size. However, the best case complexity is $O(N) + O(N)$, when all the partitions obtained are of constant size and invariant to the overall length of the time series. In practice, this brings down the average computational complexity of PDP close to $O(N)$, as we will demonstrate in evaluation section.


\subsection{Measures that qualify for PDP}\label{sec:Meas}
In this section, we discuss the relationship measures that satisfy the  three properties shown in section~\ref{sec:PDP} and qualify for PDP: \\

\noindent \textbf{Mean Square Error(MSE)} This measures distances between two time series as the mean of squares of the differences in their observations. Specifically, MSE between two time series $X$ and $Y$ on an interval $[s,e]$ is given by 
\vspace{-2em}
$$MSE(X,Y)_{[s,e]} = \frac{(\sum\limits_{t=s}^e X[t] - Y[t])^2}{l_{se}}$$



\noindent \textbf{Average Product (AP)} Average Product computes the mean of element-wise product between the observations of the two time series. Specifically, the Average Product of two time series $X$ and $Y$ in an interval $[s,e]$ is given by \vspace{-2em}
$$AP(X,Y)_{[s,e]} = \frac{\sum\limits_{t=s}^{e}X[t]*Y[t]}{l_{se}+1}$$
Both $MSE$ and $AP$ satisfies all the three properties (proofs in supplementary). Notice that the AP shares similarities with covariance measure. Specifically, for two normalized time series (zero mean, unit variance), AP in an interval measures covariance of two time series with respect to their full-length means. Further, under the assumption that the two time series are the outcome of a stationary process, the mean and variance of the two time series remain constant in any sub-interval, in which case AP would also be equal to Pearson correlation in any sub-interval.
\section{Results and Evaluation}\label{Sec:Eval}

\subsection{Data and Pre-processing}\label{SubSec:Data}
\subsubsection{Global Sea Level Pressure (SLP) Data}\label{sec:SLPData} We used monthly SLP dataset provided by NCEP/National Center for Atmospheric Research (NCAR) Reanalysis Project \cite{kistler2001ncep} which is available from 1979-2014 (36 years x 12 = 432 timestamps) at a spatial resolution of 2.5 $\times$ 2.5 degree (10512 grid points, also referred to as locations). For each time series, we followed the standard pre-processing steps followed in climate science to remove the annual seasonality and linear trends \cite{kawale2013graph}. Relationships in spatio-temporal data are preferably studied between regions (sets of spatially contiguous locations) as opposed to individual locations, since they are more reliable and stable over time. Therefore, for every location $l_i$, we grew a homogeneous and spatially contiguous region $R_i$ around it by including all the locations that showed a strong positive correlation of at least 0.85 to $l_i$, thereby resulting in 10512 regions. For every region $R_i$, we then generated its representative time series as the z-scored average time series of the constituent locations, thus getting 10512 time series of regions. 

\subsubsection{Brain fMRI Data} Functional Magnetic Resonance Image (fMRI) data measures the amount of oxygen consumed at every 2x2x2 mm voxel in the brain and is known to indicate the amount of activity occurring at any location. We used a neuro-imaging data that has been collected in a study at \cite{ramsay2013affective} on 50 subjects. In this study, participants viewed 30 sec video clips interleaved with 30 sec resting period while fMRI scans are being acquired. The temporal resolution of the scan is two seconds and the total duration of the scan was 480 seconds, thereby resulting in the length of every time series to be 240 observations. A number of fMRI pre-processing steps including motion correction, unwarping, and filtering  have been performed that are been elaborately described in \cite{ramsay2013affective}. In addition, we grouped the 2x2x2 mm voxels into 90 anatomical regions of the brain based on an Automated Anatomical Labeling Atlas \cite{tzourio2002automated}. The resultant data matrix for each subject, was of size 240 $\times$ 90. Furthermore, we observed that some of the time series show much higher variability within the intervals of resting states compared to that of video-watching states. Note that the similarity measures proposed in section~\ref{sec:Meas} are variant to scaling of the time series. Therefore, to avoid unwanted bias to either of the two states, we performed z-scoring on every resting and video-watching time interval. 



\subsection{Experimental Setup}\label{SubSec:ExptSetup}
We used measure \textit{Average Product} (AP) to capture sub-interval positive correlations in fMRI dataset, whereas for SLP dataset, we studied sub-interval negative correlations  using the measure \textit{negative Average Product} (nAP), which is exactly equal to the negative of measure $AP$.
\subsubsection{Choice of parameters}  Our problem formulation requires inputs for two parameters: $l_{min}$, the minimum length and $\tau$, the minimum strength of relationship in a sub-interval that could be selected in an SIR. The combination of two parameters determines the type of SIRs obtained in the search.
In climate science, a physical phenomenon typically shows up as a strong signal that lasts for at least six months, hence we chose $l_{min} = 6$ for SLP data. Similarly, in fMRI data, we are interested in seeking  relationships that might be prominent only during the resting or video-watching time periods. Since the length of each time period was 30 seconds, $l_{min}$ was set to 20 seconds (equivalent to 10 timestamps), a slightly smaller value to avoid noise factors. The other parameter $\tau$ was set to a high value of 1 for both experiments. Further analysis on parameter sensitivity shows robustness (see supplementary). 



\subsubsection{Selecting Candidate Pairs}
The pairs of time series with strong full-length correlations are likely to show strong correlations for most of the observations and thus, the concept of SIR has a limited relevance for such pairs. Therefore, in this paper, we limit our analysis to the pairs of time series that have weaker full-length relationships. We refer to such pairs as 'candidate pairs' to which we applied our proposed approach. For brain fMRI dataset,  all those pairs of regions that have full-length correlation weaker than 0.25 in magnitude in one of the subjects were selected as candidate pairs. Out of ${90}\choose{2}$ pairs of regions, we selected 1331 candidate pairs in this fashion. Similarly for SLP dataset, we first obtained a set of candidate pairs that have a full-length correlation weaker than 0.25 in magnitude. Due to spatial autocorrelation, a lot of pairs in $R$ are redundant with each other and need to be discarded. We treated a pair of time series $(X_1,Y_1)$  to be redundant with $(X_2,Y_2)$ if their corresponding time series are highly similar, specifically $min(AP(X_1,X_2),AP(Y_2,Y_2) \geq 0.7$. We removed redundant pairs by following a simple procedure  described in supplementary material. 

\vspace{-0.5em}
\subsection{Computational Evaluation}\label{sec:EvalComp}
\begin{figure}[t]
\centering
\includegraphics[scale=0.4]{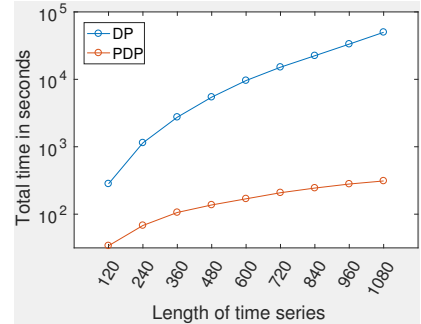}
\caption{Computational time (Y-axis) of DP and PDP for different lengths of time series (X-axis). Note that Y-axis is on logarithmic scale.}
\label{fig:PDPVsDPDiffLen}
\vspace{-1em}
\end{figure}

We evaluated evaluated PDP against DP based on their scalability, i.e. how their computational costs vary with the length of the given time series on the 14837 candidate pairs of SLP dataset. To obtain longer time series, we used global Sea Level Pressure data simulated by GFDL-CM3, a coupled physical model that provides simulations for entire $20^{th}$ century. We obtained nine time-windows of different sizes, each starting from 1901 and ending at 1910, 1920,...,1990 and for each time window, we obtained nine sets of 14837 pairs of time series. On each of these nine sets, we obtained the total computational time taken by DP and PDP  as shown in Figure~\ref{fig:PDPVsDPDiffLen}. X-axis on this figure indicates the total length of the time series (in months) for all the nine time-windows, whereas Y-axis indicates the total computational time in seconds (log scale) taken to obtain SIRs in all the candidate pairs. As expected, the computational cost of DP (blue curve) follows a quadratic function of the length of time series. However, the same for PDP (red curve) increases linearly with the length of time series. As explained earlier in section~\ref{Sec:PDPcomplx}, the complexity of PDP is $O(kN)$ in time, where $k$ is the average cost of solving the sub-problems corresponding to each partition that are further solved using DP. With all the partitions to be of a constant size, the computational time of DP for solving every sub-problem would also be constant, and therefore the resultant complexity of PDP is reduced to $O(N)$, that makes it scalable for longer time series without losing the optimality. 

\begin{figure}[h]
\centering
\begin{subfigure}[t]{0.22\textwidth}
\includegraphics[scale=0.29]{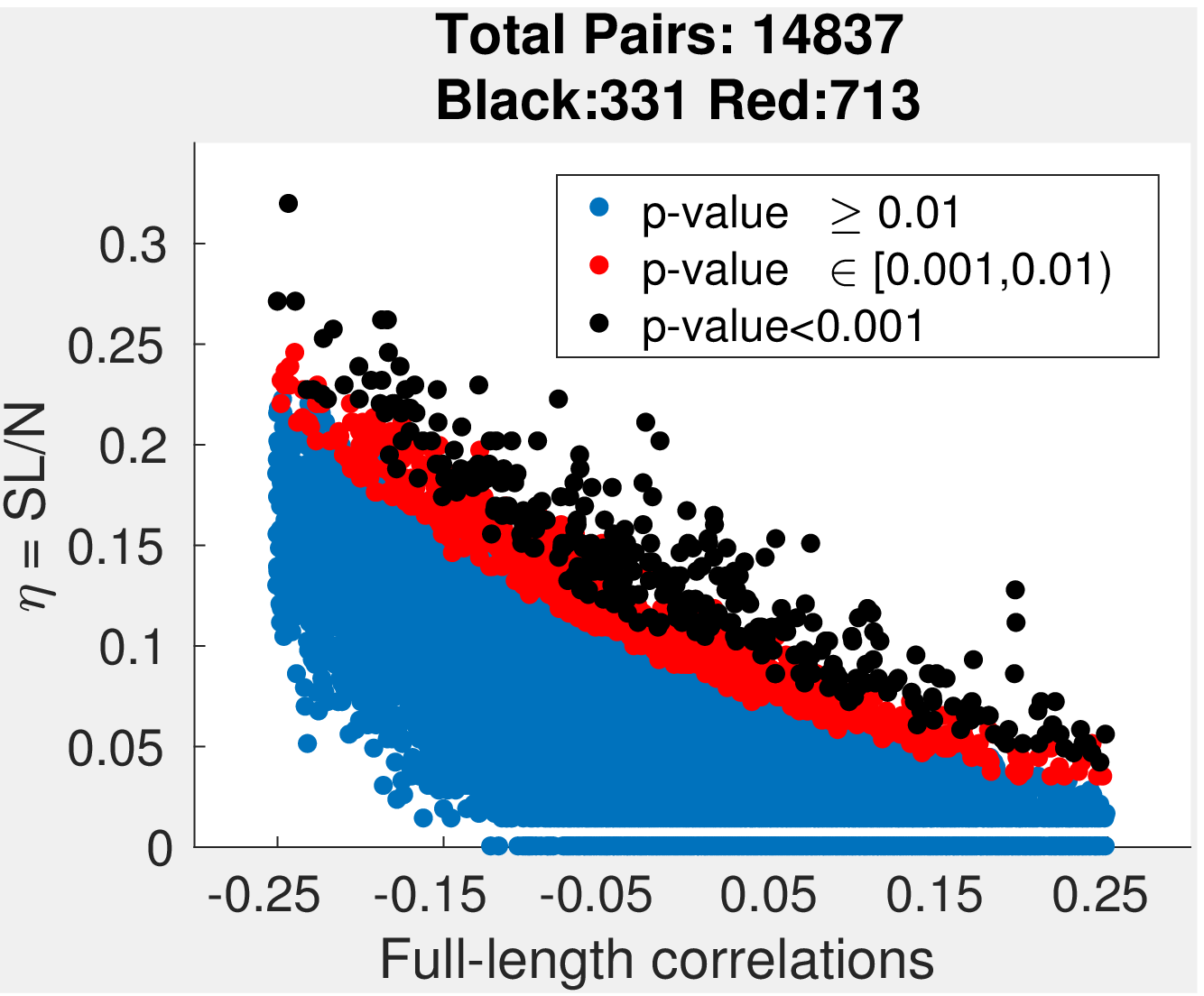}
\centering
\caption{SLP Dataset}
\label{fig:SignifSLP}
\end{subfigure}
\begin{subfigure}[t]{0.22\textwidth}
\includegraphics[scale=0.29]{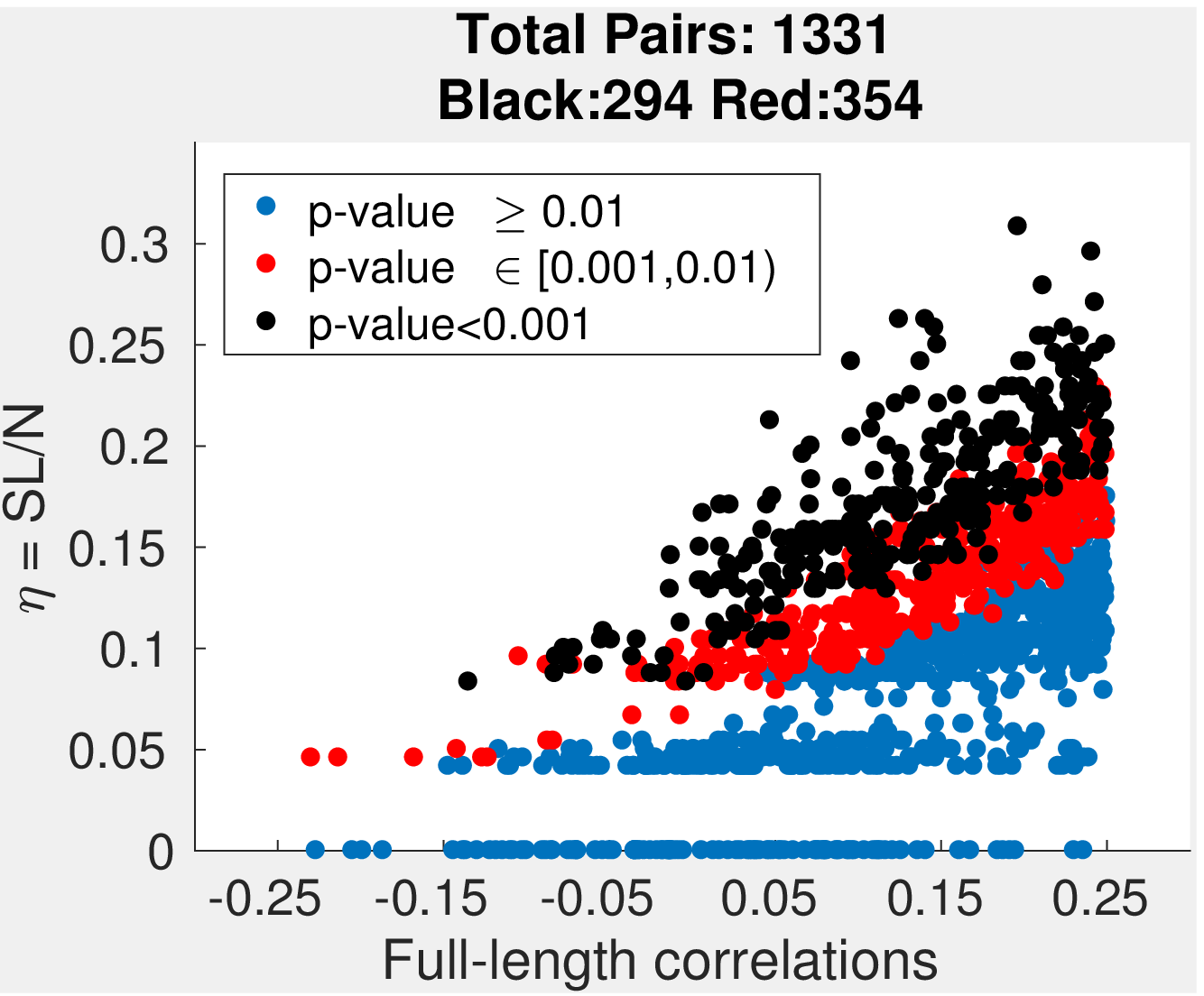}
\centering
\caption{fMRI Dataset}
\label{fig:SigniffMRI}
\end{subfigure}
\caption{All the candidate pairs with their full-length correlations on X-axis and $\eta = \frac{SL}{N}$ of their corresponding SIRs on Y-axis. Each scatter corresponds to a candidate pair and its color is  based on the p-value of its  SIR value.}
\label{fig:Signif}
\vspace{-2.0em}
\end{figure}




\begin{figure*}[!t]
\centering
\includegraphics[width=1.0\textwidth]{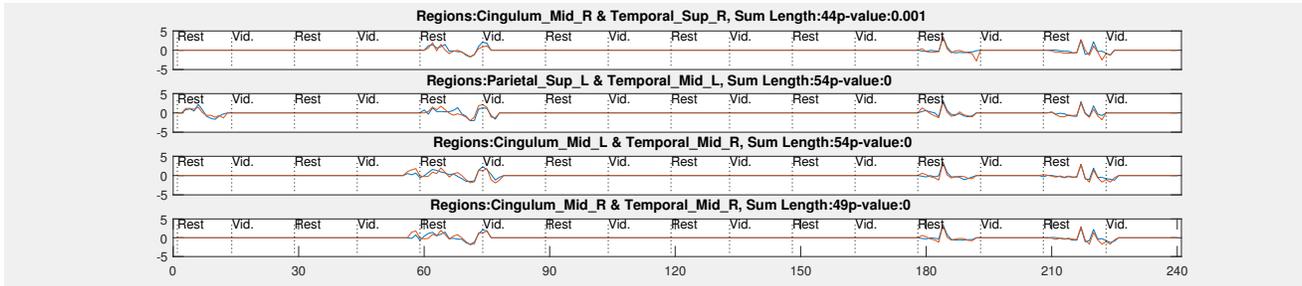}
\caption{A set of four SIRs that synchronize with each other in more than 40 timestamps for 10 subjects. Above shown are the time series from the scan of one of the subjects.}
\label{fig:assocfMRI}
\vspace{-1em}
\end{figure*}

\subsection{Evaluation of Discovered SIRs}\label{sec:signif}
The lack of ground truth prohibits us to do a systematic domain evaluation on all the candidate pairs. Therefore, we evaluate the validity of discovered SIRs by analyzing the statistical significance of the sum-lengths of the obtained SIRs. We followed a randomization-based approach that addresses the question: Can the sum-length of the given SIR between $T_1$ and $T_2$ be achieved by replacing one of the two time series with a random time series whose full-length correlation with the fixed time series is preserved? If this happens for a very few random time series, then the SIR obtained between $T_1$ and $T_2$ can potentially considered to be statistically significant. The approach was setup in the following fashion: For each of the discovered SIR in a pair of time series $T_1$ and $T_2$, without loss of generality we replaced time series $T_1$ by a random time series $T_{rand}$ and recomputed the sum-length of the optimal set corresponding to the SIR formed between $T_{rand}$ and $T_2$. The random time series is generated such that its full-length correlation with the other fixed time series $T_2$ is preserved, i.e. $corr(T_{rand},T_2) = corr(T_{1},T_2)$. This process was repeated 1000 times and a distribution of sum-lengths was obtained that was used to compute the p-value of the original sum-length of the optimal set corresponding to the SIR between $T_1$ and $T_2$. Specifically, the p-value was computed as the fraction of the random time series that formed an SIR of higher sum-lengths with $T_2$. 


Using above procedure, we evaluated the statistical significance of all of the SIRs discovered in SLP and brain fMRI datasets. Figure~\ref{fig:Signif} shows two scatter plots between strength of full-length correlations (X-axis) and the $\frac{SL}{N}$, the normalized sum-length of the SIRs in all the candidate pairs of SLP and brain fMRI dataset. To emphasize the statistical significance at different levels, we depict each pair with either blue (p-value $\geq 0.01$) or red (p-value $\in [0.01,0.001]$) or black (p-value $< 0.001$) scatter in the Figure~\ref{fig:Signif}. The red and black scatter together constitute total of $1044$ statistically significant SIRs. Further, there are at least $331$ (black)  pairs of highly significant pair with p-value $<0.001$. 
Similarly, Figure~\ref{fig:SigniffMRI} shows 648 out of 1331 candidate pairs that were found to be statistically significant in fMRI dataset. Notice that the significance of an SIR depends on both its sum-length and the full-length correlation of the pair. In fMRI (SLP) dataset, pairs with weaker positive (negative) full-length correlations showed significant SIRs at smaller sum-lengths compared to the pairs with stronger positive (negative) full-length correlations. This is expected because in a pair with positive (negative) full-length correlations, the timestamps with high negative (positive) AP are expected to be rarer, and finding them together as long sub-intervals is furthermore unlikely, compared to the cases where such timestamps are more in number.




\begin{figure}[!h]
\centering
\includegraphics[scale=0.4]{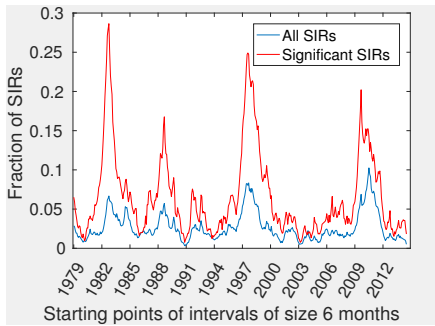}
\centering
\caption{For every interval of size 6 months, the plot indicates the proportion of i) 1044 significant SIRs (red curve) and ii) all 14837 candidate SIRs (blue curve) that included given interval.}\label{fig:EventSLP}
\vspace{-1.5em}
\end{figure}

\vspace{-0.7em}
\subsection{Applications and Domain Insights}\label{Sec:DomInsights}
While SIRs are interesting by themselves in finding non-trivial relationships that are otherwise difficult to be revealed using full-length relationship measures, studying them collectively in a time series data could provide further insights about the data. Here we discuss two potential approaches for their collective analysis.

\vspace{-0.7em}
\subsubsection{Finding Anomalous Intervals} A potential application of this work could be to detect anomalous time intervals that experience unusually high number of relationships. Specifically, for every interval $[s,e]$, one can obtain a score that indicates the proportion of candidate pairs that were 'active' during entire $[s,e]$. Intervals included in unusually high number of SIRs could potentially indicate the occurrence of a special event. Applying this idea to SIRs of SLP dataset, we obtained the scores for all possible intervals of size 6 months as shown in Figure~\ref{fig:EventSLP}. It can be seen that the scores are anomalously high for the intervals 1982 Sept-83 Mar, 1988 Sept -89 Mar, 1997 Aug -98 Feb, and 2009 Sept -10 Mar. All of the above intervals are known to have experienced the strongest el-nino and la-nina events since 1979 \cite{ensoyears}. During these events, climate behaves quite differently compared to the general climatology. New wave patterns emerge that synchronize regions with each other that are otherwise unrelated to each other. 

\vspace{-0.7em}
\subsubsection{Discovery of Associated SIRs}
Another potential utility of this work could be to explore interesting connections that could exist between different pairs of regions that showed significant SIRs. For instance, there could exist certain candidate pairs that show SIRs simultaneously on multiple occasions. We refer to such sets as "Associated SIRs" due to their strong association in multiple intervals of time. Such associated SIRs can be found by applying frequent pattern mining to a given pool of pairs with SIRs, where each pair is as an item and each timestamp is a transaction. 

Using this approach, we discovered a set of four SIRs (shown in Figure~\ref{fig:assocfMRI}) that share highly similar intervals in the block-design fMRI dataset where the subject is exposed to interleaved video and rest periods. It is interesting to note that these SIRs are formed among parietal and temporal regions that are known to be activated by the visual and auditory stimulus \cite{downar2001effect}, respectively. It is also worth noting that one of the brain regions Cingulum participates in three of the four SIRs, which is known to be involved in integrating memory with other regions of the brain \cite{de2005fmri}. While the regions involved in these SIRs are related to the video task at hand, it is surprising that these SIRs mostly capture intervals from the `rest' segments, and not `video' segments that is known to simultaneously activate the involved regions. While these observations need to be validated by domain scientists, this example underscores the utility of the proposed approach for discovering SIRs in brain fMRI data. 
 


\vspace{-0.5em}
\section{Conclusion}\label{Sec:Conc}

In this paper, we defined a notion of sub-interval relationship to capture interactions between two time series that are intermittent in nature and are prominent only in certain sub-intervals of time. We proposed a novel approach to find most interesting SIR in a pair of time series that guarantees to find the optimal SIR. We also demonstrated the scalability of our approach and showed it to be efficient by an order of $N$ compared to standard dynamic programming approach in practice. We further exhibited the utility of  SIR in two real-world applications: climate and neuroscience to obtain useful domain insights. 

\vspace{-1em}
\footnotesize
\bibliographystyle{IEEEtran}
\bibliography{IEEEabrv}

\end{document}